# Multivariate feature selection and autoencoder embeddings of ovarian cancer clinical and genetic data

Luis Bote-Curiel [a], Sergio Ruiz-Llorente [b], Sergio Muñoz-Romero [a], Mónica Yagüe-Fernández [b], Arantzazu Barquín [b], Jesús García-Donas [b], José Luis Rojo-Álvarez [a,*]

[a] Department of Signal Theory and Communications, Universidad Rey Juan Carlos, 28942 Fuenlabrada (Madrid), Spain
[b] Unit of Gynecological, Genitourinary and Skin Tumors, Hospital HM Sanchinarro, Fundación Investigación HM Hospitales, 28050 Madrid, Spain

*Keywords:*
Autoencoders
Clinical data
Feature extraction
Feature selection
Genetic data
Ovarian cancer
Platinum-resistant
Platinum-sensitive

A B S T R A C T

Although certain genetic alterations have been defined as predictive and prognostic biomarkers in the context of ovarian cancer (OC), data science methods represent alternative approaches to identify novel correlations and define relevant markers in these gynecological tumors. Considering this potential, our work focused both on clinical and genomic data information collected from patients with OC to identify relationships between clinical and genetic factors and disease progression-related variables. For this aim, we proposed two analyses: (1) a nonlinear exploration of an OC dataset using autoencoders, a type of neural network that can be used as a feature extraction tool to represent a dataset in 3-dimensional latent space, so that we could assess whether there are intrinsic or natural nonlinear separability patterns between disease progression groups (in our case, platinum-sensitive and platinum-resistant groups); and (2) the identification of relevant variable relationships by means of an adaptation of the informative variable identifier (IVI), a feature selection method that labels each input feature as informative or noisy with respect to the task at hand, identifies the relationships among features, and builds a ranking of features, allowing us to study which input features and relationships may be most informative for the OC disease progression classification to define new biomarkers involved in disease progression. Our interest has been in clinical and genetic factors and in the combination of clinical features and genetic profile. Results with autoencoders suggest a pattern of separability between disease progression groups in the clinical part and for the combination of genes and clinical features of the OC dataset, that is increased via supervised fine tuning. In the genetic part, this pattern of separability is not observed, but it is more defined when a supervised fine tuning is performed. Results of the IVI-mediated feature selection method show significance for relevant clinical variables (such as type of surgery and neoadjuvant chemotherapy), some mutation genes, and low-risk genetic features. These results highlight the efficacy of the considered approaches to better understand the clinical course of OC.

## 1. Introduction

Ovarian cancer (OC) constitutes one of the most serious health problems in our society, being the second most common gynecological neoplasm with an estimated annual incidence of 225 000 women worldwide (Moschetta, George, Kaye, & Banerjee, 2016). Furthermore, OC is the fifth leading cause of cancer-associated mortality and is the gynecological tumor with the worst prognosis (140 000 deaths per year), with a 5-year overall survival close to 15% (Stewart, Ralyea, & Lockwood, 2019). Various basic and clinical studies have confirmed the role of inactivating mutations in the *BRCA1* and *BRCA2* genes, homologous recombination (HR) pathway deficiencies, and certain alteration profiles (HR-related mutational signature 3 or specific copy number signatures) as prognostic and predictive markers of ovarian carcinomas. However, a certain percentage of patients not showing such alterations still obtain clinical benefit from OC standard therapies (Macintyre et al., 2018).

Data science methods are currently widely applied in the context of OC. Methods such as machine learning and, more specifically, deep learning, are being applied to the clinical research of diagnostic imaging procedures, either histological, ultrasound, or computed tomography images, to better define the anatomopathological diagnosis or to infer the treatment response and the malignant properties of OC tumors under study (Zhu, Xie, Han, & Guo, 2020). However, there are limited works regarding the machine-learning-mediated integration of clinical information and molecular data obtained from -omic platforms, which have mostly focused on transcriptomics and proteomics profiling. For instance, one recent work used neural networks to identify a subset of immune-related proteins and catabolic pathways correlating with diagnosis-related clinical variables (tumor grade) (Yu et al., 2020). Another work has also shown potential for efficient diagnosis of epithelial ovarian carcinomas using a neural network and serum microRNA (Elias et al., 2017).

* Corresponding author.
*E-mail addresses:* luis.bote@urjc.es (L. Bote-Curiel), sruizllorente@hmhospitales.com (S. Ruiz-Llorente), sergio.munoz@urjc.es (S. Muñoz-Romero), myaguefernandez@hmhospitales.com (M. Yagüe-Fernández), abarquin@hmhospitales.com (A. Barquín), jgarciadonas@hmhospitales.com (J. García-Donas), joseluis.rojo@urjc.es (J.L. Rojo-Álvarez).

Feature selection methods are another group of Data Science techniques that have been used in the context of OC, and they aim to highlight which are the input features that have the strongest relationship with the target features, thus providing an indication of the importance of each of them. Depending on the degree of interpretability of the feature selection methods (Muñoz-Romero, Gorostiaga, Soguero-Ruiz, Mora-Jiménez, & Rojo-Álvarez, 2020), some of them can also identify relationships among features that could be especially informative for medical researchers, in a way that these methods can indicate which features are relevant to a particular disease and how those features interact with each other. Typically, feature selection methods are classified into filter, embedded, and wrapper methods, depending on their relationship with the learning method. Filter methods do not depend on any learning method and require less computational time, while wrapper and embedded methods both require a learning method to perform feature selection (Saeys, Inza, & Larrañaga, 2007). Nowadays, some of the most widely used feature selection methods are still classical algorithms such as the *Relief* method (Kira & Rendell, 1992; Kononenko, 1994) or those algorithms and criteria based on mutual information, including conditional mutual information (CMI) (Brown, Pocock, Zhao, & Luján, 2012), mutual information feature selection (MIFS) (Battiti, 1994), joint mutual information (JMI) (Yang & Moody, 1999), minimum redundancy maximum relevance (MRMR) (Ding & Peng, 2005; Peng, Long, & Ding, 2005), mutual information maximization (MIM) (Lewis, 1992), conditional mutual information maximization (CMIM) (Fleuret, 2004), interaction capping (ICAP) (Jakulin, 2005), double input symmetrical relevance (DISR) (Meyer & Bontempi, 2006), conditional infomax feature extraction (CIFE) (Lin & Tang, 2006), or conditional redundancy (CondRed) (Brown et al., 2012). However, these classic feature selection methods, algorithms, and criteria have been rediscovered countless times and many versions of each have been implemented. For example, the *Relief* method has even generated a family of filter-style feature selection algorithms called Relief-based algorithms (RBAs) (Urbanowicz, Meeker, La Cava, Olson, & Moore, 2018). For algorithms and criteria based on mutual information, the literature is vast (Vergara & Estévez, 2014). For example, we can highlight new variants of MRMR, such as the Fast-MRMR algorithm (Ramírez-Gallego et al., 2017) or the temporal minimum redundancy maximum relevance (TMRMR) feature selection approach (Radovic, Ghalwash, Filipovic, & Obradovic, 2017). For CMI, we can point out a variation called conditional mutual information based feature selection considering Interaction (CMIFSI) (Liang, Hou, Luan, & Huang, 2019). And for JMI, we also can mention joint mutual information maximisation (JMIM) and normalised joint mutual information maximization (NJMIM) algorithms (Bennasar, Hicks, & Setchi, 2015). Furthermore, in the last years, feature selection methods have been widely applied in medicine, for example, in medical imaging, biomedical signal processing, or genomic data. A comprehensive review of feature selection methods in medical applications can be found in Remeseiro and Bolon-Canedo (2019), and more specifically on genetic data, a review of feature selection can be found in Tadist, Najah, Nikolov, Mrabti, and Zahi (2019). Also, feature selection methods have been applied, for example, to the discovery of new biomarkers of diseases. Several works on this topic have been applied to OC, but they are mostly focused on proteomics (Alipoor, Khani Parashkoh, & Haddadnia, 2010) and serum biomarkers (Song et al., 2018), rather than on the integration of clinical and molecular data.

Taking this scenario into account, our research group developed two previous analysis of OC data mixing clinical and genetic features (Bote-Curiel et al., 2021a, 2021b). In these preceding works, we developed a bootstrap framework for two different types of analysis. In the first one, we performed an initial univariate analysis of features of different types that provides useful knowledge about the quality of an OC dataset, and a first set of variables was identified as potentially relevant for OC disease progression classification (Bote-Curiel et al., 2021a). In the second one, we conducted non-supervised linear multivariate analysis that represents a useful feature extraction tool for variables with different nature (namely numerical, categorical, and text) and provided information about intrinsic patterns in the OC dataset under study (Bote-Curiel et al., 2021b). However, in these two works we had not explored the heterogeneous OC data either in terms of supervised nonlinear multivariate data models, or by using convenient feature selection methods including relations among said heterogeneous features. These two approaches represent two relevant and subsequent stages in systems for machine learning analysis and they are crucial for the final performance and interpretability of data-driven model predictions, and they deserve their own entity and attention. It is for these reasons that we are specifically addressing them in the present work.

Therefore, we focus here on both extensive clinical information collected from patients diagnosed with OC and genomic data obtained through whole exome or targeted sequencing, with the aim of identifying relationships between clinical and genetic factors and OC disease progression-related variables. For this purpose, we propose two analyses. The first analysis is a nonlinear exploration of an OC dataset through autoencoders. An autoencoder is a type of neural network that can be used as a feature extraction tool for distilling relevant information from data (Martínez-Ramón, Gupta, Rojo-Álvarez, & Christodoulou, 2020). Specifically, we use autoencoders to compress the data into a 3-dimensional latent space so that we could assess whether the dataset has intrinsic or natural separability patterns between disease progression groups (platinum-based therapy sensitive vs. resistant groups). The second analysis consists of using an adaptation of the informative variable identifier (IVI) algorithm (Muñoz-Romero et al., 2020), a feature selection tool that labels each input feature as informative or noisy with respect to the task at hand, identifies the relationships among features, and builds a ranking of features. In our case, this approach enables us to study which input features and relationships may be most informative for OC disease progression classification, hence being able to discover new biomarkers of the disease.

This paper is organized as follows. In Section 2, we describe the OC dataset. In Section 3, we present the data analysis methods. Then, the results of applying these analysis methods to the OC dataset are provided in Section 4. Discussion and conclusions are established in Section 5. Finally, in Appendix, we present simple examples of the use of the described analysis methods applied to synthetic data that can be useful to the interested reader.

**2. Dataset description**

The OC dataset that we used in this work is part of the *BRCAness* initiative from the Innovation Oncology Laboratory of the Gynecological, Genitourinary, and Skin Cancer Department, at Clara Campal Comprehensive Cancer Center (Madrid, Spain). This department has conducted a multicenter observational study focused on the identification of biomarkers with potential impact in clinical practice. In this observational study, which is supported by 12 national health care institutions, approximately 300 patients with OC have been included thus far. Inclusion criteria include both age (>18 years old) and disease status (Stage IC or superior). Among these patients, 54 were molecularly characterized by means of next generation sequencing (NGS), either

**Table 1**
Variables in the genetic part of the OC dataset.

| Name | Description |
|---|---|
| HGNC_Symbol | Gene code regarding the Human Genome Nomenclature Committee. |
| Chr | Chromosome name. |
| Genetic_Change | Genetic change in the variant allele with respect to the reference allele. |
| Genotype | Genotype name. |
| VarDepth | Number of times that the variant allele has been read in a specific region. |
| Conservation_Score | Score of the conservation in an evolutionary sense. |
| Grantham_Distance | Distance between the variant amino acid and the reference amino acid in an evolutionary sense. |
| Condel_Prediction | Condel tool prediction of pathogenicity. |
| Condel_Prediction_Score | Score of the Condel tool degree of pathogenicity. |
| Sift_Prediction | Sift tool prediction of pathogenicity. |
| Sift_Prediction_Score | Score of the Sift tool degree of pathogenicity. |
| PolyPhen_Prediction | PolyPhen tool prediction of pathogenicity. |
| PolyPhen_Prediction_Score | Score of the PolyPhen tool degree of pathogenicity. |
| Impact | Prediction of pathogenicity. |
| Amino_Acids | Variant amino acid. |
| PFI | Platinum-free interval, which represents the time from the last cycle of chemotherapy treatment to the evidence of disease progression. |

**Table 2**
Variables in the clinical part of the OC dataset.

| Name | Description |
|---|---|
| Oncological_History | Whether or not the patient has had any type of cancer. |
| Gynecological_Family_History | Whether or not the patient's family has had a member with gynecological cancer. |
| Status_BRCA | Whether or not there is a mutation in the BRCA1 and BRCA2 genes. |
| Age_at_Diagnosis | Patient's age at cancer diagnosis. |
| Anatomical_Location | Anatomical location of the patient's tumor. |
| Histology_1st_Component | Type of ovarian cancer the patient has. |
| Grade | Grade of the ovarian tumor the patient has. |
| Perineural_Vascular_Invasion | Existence or not of vascular or perineural invasion in the patient's tumor. |
| Stage | Stage of the ovarian tumor the patient has. |
| Surgery | Type of surgery performed on the patient: primary or interval. |
| HIPEC_in_Surgery | Whether or not the hyperthermic intraperitoneal chemotherapy treatment has been applied to the patient. |
| Type_of_Primary_Surgery | Type of primary surgery performed on the patient (if applicable): complete resection of the tumor or not. |
| Neoadjuvance | Whether or not the chemotherapy treatment prior to primary surgery has been applied to the patient. |
| Response_of_Neoadjuvance | How the patient has responded to neoadjuvant chemotherapy (if applicable). |
| Attitude_of_Interval_Surgery | Whether or not an interval surgery is performed on the patient after neoadjuvant chemotherapy (if applicable). |
| Type_of_Interval_Surgery | Type of interval surgery performed on the patient (if applicable): complete resection of the tumor or not. |
| Adjuvance | Whether or not the chemotherapy treatment after the primary surgery has been applied to the patient. |
| Response_of_Adjuvance | How the patient has responded to adjuvant chemotherapy (if applicable). |
| PFS | Progression-free survival, which represents the time from the first date of chemotherapy treatment to the evidence of disease progression. |
| PFI | Platinum-free interval, which represents the time from the last date of chemotherapy treatment to the evidence of disease progression. |
| OS | Overall survival, which represents the duration of patient survival. |
| Bevacizumab_Maintenance | Whether or not the antiangiogenic treatment has been applied to the patient. |

with whole-exome sequencing (WES) or with predesigned targeted gene panels (Onco80).

WES profiling (SureSelect Human All Exon V6) was performed in 20 patients. Sequencing was performed using genomic DNA extracted from either formalin-fixed paraffin embedded tumoral tissue or peripheral blood to ultimately define somatic and germinal variants. Subsequently, 34 patients showing intermediate degree of response to platinum agents were screened using an Onco80 predesigned mutational panel that enables identification of genetic alterations in 80 *loci* widely associated with cancer development. Therefore, the genetic part of the dataset consists of 7 077 observations corresponding to genetic alterations for the studied patients. The most relevant genetic variables included genetically altered *loci*, aminoacid substitution, pathogenicity scores (Grantham's distance, conservation scores according to several *in silico* programs), and the resultant genetic changes. The names and descriptions of the genetic variables used in this work are presented in Table 1.

The clinical information of each patient was also collected including age at diagnosis, personal or familial antecedents, *BRCA* and *TP53* status, histological subtype, grade and stage of the disease at diagnosis, anatomical location, presence of perineural or vascular invasion, CA-125 biomarker evaluation, surgical procedures, information related to the different treatment lines prescribed for each patient (number of cycles, doses, associated toxicities, grade of response and relapses), and date of the last clinical follow up or exitus. Other important clinical features, such as overall survival (OS), progression-free survival (PFS), and platinum-free interval (PFI), were also included. In total, the clinical part of the dataset consists of 54 entries (one for each patient) and 106 clinical variables. In Table 2, we present the names of the clinical variables and their descriptions.

## 3. Data analysis

As previously commented, in one of the earlier works, we used linear feature extraction methods to create new and smaller sets of variables that captured most of the useful information in an OC dataset (Bote-Curiel et al., 2021b). In this way, by compressing the OC dataset into a 3-dimensional latent space, we could assess whether the dataset had intrinsic or natural separability patterns between disease progression groups. However, these methods were based on linear assumptions of the data relationships, which could be invalid. Therefore, in this work, we extend the exploration of the OC dataset using autoencoders, a nonlinear feature extraction neural network that enables us to assess the intrinsic separability patterns between disease progression groups without making any prior assumption of the data. In addition to this type of information provided by feature extraction methods, we are also interested in identifying which features of the OC dataset are most relevant for disease progression group prediction so that they can be used as possible biomarkers. For this task, we used a feature selection method that extracts the most informative features in a dataset and identifies the relationships among them according to a target variable in a supervised machine learning scheme.

## 3.1. Autoencoders

An autoencoder is a flexible nonlinear feature extraction method that is able to create a low-dimensional set of variables that represents most of the useful information in a dataset. An autoencoder consists of a simple neural network whose goal is to transform inputs into outputs that contain the same information as the inputs with the least possible distortion. Therefore, an autoencoder is a neural network that is trained to replicate its input to its output (Hinton & Salakhutdinov, 2006).

Various kinds of autoencoders exist; our first interest is in the undercomplete autoencoders, which by design reduce data dimensions utilizing a bottleneck architecture that first turns a high-dimensional input into a latent low-dimensional code and then perform a reconstruction of the input using this latent code (Martínez-Ramón et al., 2020). If the latent code is 3-dimensional, we can visually and quantitatively assess whether the dataset has intrinsic or natural separability patterns.

The bottleneck architecture of an autoencoder consists of two substructures, namely, an encoder and a decoder. Though both substructures can have several layers, we focus on the single-layer case (shallow autoencoder). Such an encoder maps a high-dimensional input to latent low-dimensional code as

$$\mathbf{h}_i = f(\mathbf{x}_i) = \phi(\mathbf{W}_e \mathbf{x}_i + \mathbf{b}_e), \quad (1)$$

where $f(\cdot)$ is the transformation from the input space to the latent space, $\mathbf{h}_i \in \mathbb{R}^d$ is the code in the latent space corresponding to input $\mathbf{x}_i \in \mathbb{R}^D$, and $\phi(\cdot)$ denotes a nonlinear activation function.

The decoder again transforms the code in the latent space to the input space as

$$\mathbf{o}_i = g(\mathbf{h}_i) = \varphi(\mathbf{W}_d \mathbf{h}_i + \mathbf{b}_d), \quad (2)$$

where $g(\cdot)$ is the transformation from the latent space to the input space, $\varphi(\cdot)$ is a nonlinear activation function, and $\mathbf{o}_i$ is the estimated output. The problem is essentially to estimate the functions $f(\cdot)$ and $g(\cdot)$ that make $\mathbf{o}_i$ as similar as possible to $\mathbf{x}_i$. This goal is achieved by calculating weights and biases $\mathbf{W}_e$, $\mathbf{W}_d$, $\mathbf{b}_e$, and $\mathbf{b}_d$ through an iterative optimization process with an appropriate loss function. This is called the training process.

The use of nonlinear activation functions allows the neural network to model output features that vary non-linearly with respect to its inputs features. Although there are many types of activation functions, the most commonly used are the rectified linear (ReLU) or one of its variations (leaky ReLU or parametric ReLU), which are the default activation functions recommended for using with most neural networks hidden units. Previously, the default activation function in hidden units was the hyperbolic tangent activation function, but it has been replaced by ReLU because, in general, the performance is better in a variety of practical applications. In binary classification problems, the sigmoid activation function is used in the output units since the output is interpreted as a binomial probability distribution. In multiclass classification problems, the sigmoid activation function is being replaced by the softmax activation function, because the output is then interpreted as a multinomial probability distribution (Goodfellow, Bengio, & Courville, 2016).

Autoencoders have traditionally been used as nonlinear dimensionality reduction tools in feature extraction or for dataset visualization when the code has 2 or 3 dimensions. To illustrate the functioning of this method, we present a simple example with a synthetic dataset in Appendix.

In addition to undercomplete autoencoders, other types exist. Denoising autoencoders aim to reconstruct a partially corrupted input rather than simply copying the input; sparse autoencoders are simply autoencoders whose training loss function involves a sparsity penalty; variational autoencoders are generative methods (Goodfellow et al., 2016).

## 3.2. Informative variable identifier

We are interested here in finding the most relevant features of our dataset for prediction and for interpretability purposes. For the latter, the relationships among these features are specially important for the clinicians. Feature selection methods perform the task of extracting the most informative features in a dataset, but only some of the existing methods provide us with explicitly identified relationships among features. Our group has recently presented the IVI algorithm (Muñoz-Romero et al., 2020), a computationally competitive feature selection method for both selecting features and identifying relationships among them. Specifically, given a classification task and input features, the IVI algorithm: (1) categorizes each input feature as informative or non-informative with respect to the classification task under study; (2) detects the relationships among informative input features; and (3) ranks informative input features by importance. An explanation of this method is presented next.

IVI starts with an input data matrix, $\mathbf{X} \in \mathbb{R}^{L \times N}$, with $L$ observations, $\mathbf{x}_l$, each with $N$ features. The focus is on classification with a binary output variable, $\mathbf{y} \in \mathbb{R}^L$, with $y_l \in \{-1, +1\}$ for $l = 1, \ldots, L$. The method is designed based on the hypothesis that weights, $\mathbf{w}$, that are generated by a linear classifier, $y_l = \mathbf{w} \cdot \mathbf{x}_l + b$, are able to explain the relationship between each input feature and the output, as well as the relationships among the input features. Furthermore, IVI uses the statistical properties of the weights to obtain information about which input features are informative or not, and it also detects redundancies.

The statistical properties of the weights are based on their probability density function (*pdf*), denoted as $f_\mathbf{w}(\mathbf{w})$, which is a multivariate function that provides complete statistical knowledge about the weights. Therefore, the initial step involves calculating $f_\mathbf{w}(\mathbf{w})$. However, this *pdf* is often difficult to calculate, so IVI instead works with the marginal *pdf* of the weights, denoted as $f_{w_n}(w_n)$ for the $n$th input variable.

Bootstrap resampling, which can estimate the empirical distribution of any statistics that can be calculated computationally (Efron & Tibshirani, 1986), is used to estimate $f_{w_n}(w_n)$ for each $n$th input variable. In this case, IVI uses the input data matrix, $\mathbf{X}$, and the output variable, $\mathbf{y}$, to generate the $b$th resample $\mathbf{X}^*_{(b)}$ and $\mathbf{y}^*_{(b)}$. These results are then used to estimate the weights, $\mathbf{w}^*_{(b)}$. The estimated marginal *pdf* of each weight, $f^*_{w_n}(w_n)$, is then calculated by repeating this process $B$ times.

By means of the above steps, IVI identifies the informative variables. To detect these informative variables, the confidence interval (CI) of $f^*_{w_n}(w_n)$ for each weight is analyzed, where it is assumed that informative features exhibit non-zero overlapping CIs. Therefore, features with this criterion are informative variables and are called *relevant variables*. However, it is likely that not all the informative variables are included in the set of relevant variables because of feature redundancy. Thus, there is interest in identifying the redundant variables that could also be informative.

To detect redundancy, the Pearson correlation of weights across replicates is used as follows,

$$\rho^*_{m,n} = corr(\mathbf{w}^*_m, \mathbf{w}^*_n), \quad (3)$$

where $\mathbf{w}^*_m$ and $\mathbf{w}^*_n$ are the vectors of the estimated weights $m$ and $n$, respectively, and *corr* denotes the correlation coefficient. IVI categorizes $m$ and $n$ features as redundant when the absolute value of their correlation coefficient, $|\rho^*_{m,n}|$, exceeds a $\rho_{th}$ threshold (that is, $|\rho^*_{m,n}| > \rho_{th}$). Redundant features can be clustered to create disjoint subsets such that a feature is assigned to a disjoint subset if it is redundant with respect to one o more features within the subset. With this, informative features can be identified by selecting the disjoint subsets with at least one feature identified as relevant. Thus, subsets without any relevant feature are discarded, and all their features can be considered as non-informative, while features in subsets with at least one relevant variable are considered as informative.

The above process yields a list of informative variables and also indicates redundancy relationships. However, users may also be interested in a subset of the informative variables. Therefore, IVI includes a ranking of variables. For each informative variable, $n$, an importance score is calculated as the absolute value of the mean across $\mathbf{w}_n^*$ divided by the square of the range of the 95% CI of $\mathbf{w}_n^*$, that is,

$$Imp_n = \frac{|mean(\mathbf{w}_n^*)|}{(w_n^{h*} - w_n^{l*})^2}. \quad (4)$$

Then, each informative feature is ordered by descending importance score to create the ranking.

Although IVI was designed to use linear classifier weights, it can be implemented with any machine learning classifier that computes feature weights. For example, the approach proposed in Muñoz-Romero et al. (2020) for this purpose was the covariance multiplication estimator (CME), which is a method that produces a nonlinear transformation of the input feature space into a weight space intended to be computationally competitive with the linear algorithms. A simple explanation of the CME is as follows. Given $\mathbf{X}$ and $\mathbf{y}$, sample covariance matrices $\mathbf{C_{XX}}$ and $\mathbf{C_{Xy}}$ are calculated. Given these matrices, the feature weights, $\mathbf{w}$, are defined as follows:

$$\mathbf{w} = (sign(\mathbf{C_{XX}}))^{(g-1)} \odot (\mathbf{C_{XX}})^{(g)} \mathbf{C_{Xy}}, \quad (5)$$

where $\odot$ is the Hadamard product (element-wise product) and $g$ denotes the integer exponent of the element-wise power, which represents a trade-off between $\mathbf{C_{XX}}$ or $\mathbf{C_{Xy}}$.

### 3.3. Adaptation of IVI

IVI is suitable to find the most informative features and the relationships among features in a dataset and can be used as a feature selection method and as an interpretability technique. However, in this work, we propose several modifications to improve both IVI and CME to make them more stable. For CME, this method works better when it handles metric features, but when there is a set of mixed metric and binary features, it can fail to detect relationships between metric and binary features and among binary features. Therefore, given $\mathbf{X}$ and $\mathbf{y}$, we propose to use the sample correlation matrices, $\mathbf{R_{XX}}$ and $\mathbf{R_{Xy}}$, to calculate the feature weights, $\mathbf{w}$, as follows:

$$\mathbf{w} = \mathbf{R_{XX}} \mathbf{R_{Xy}}. \quad (6)$$

With this simple and effective adaptation, CME can detect the relationships among metric features, among binary features, and between metric and binary features.

Other than this, IVI performs properly with large datasets but it can be unstable with small datasets, since in different runs the algorithm returns different solutions. To solve this, we propose to use IVI in a probabilistic manner; that is, instead of calculating the IVI once, we run and calculate the IVI $M$ times. Thus, we obtain $M$ sets of informative features with their respective relationships, so we can filter the features according to the number of times they appear in those $M$ repetitions. Specifically, given the $M$ sets of relevant informative variables, an operative threshold can consists of keeping the features that appear 90% of the time, consistent with a high-demand criterion such as that proposed by IVI. Additionally, given a relationship between two features, we have a measure of that relationship using the correlation between them. Thus, given the $M$ sets of relationships, we can calculate a global measure of each relationship as the sum of the correlation in the $M$ sets. In this way, we can filter redundant informative variables using the maximum of the derivative of the cumulative sum of the global measures in descending order, in the same way that it is calculated in IVI. A synthetic example of IVI with the CME method with these modifications is presented in Appendix.

## 4. Experiments and results

In this section, we apply the methods presented in the previous section to analyze the OC dataset. We first transform each observation of the OC dataset to a 3-dimensional latent space using autoencoders to find visual patterns in the data. Then, we use an adaptation of IVI to study the most relevant features in the OC dataset for OC disease progression classification and to assess the relationships among these features.

### 4.1. Autoencoders in the OC dataset

To understand the relations of clinical and genetic data with OC progression, we explore the OC dataset presented in Section 2 with autoencoders, which are described in Section 3. In this case, autoencoders can be used as a feature extraction tool to reduce the dimensionality of the observations to a 3-dimensional latent space. Specifically, we transform observations in the OC dataset to a 3-dimensional latent space so that we can visually assess the intrinsic or natural separability patterns between OC disease progression groups. In particular, these groups are based on an indicator of disease progression called PFI, which is defined as the time (in months) between the last cycle of platinum and evidence of disease progression (Pujade-Lauraine & Combe, 2016). In this setting, depending on the length of platinum drug sensitivity, patients were categorized into the platinum-resistant group ($< 6$ months) or platinum-sensitive group ($> 6$ months). Therefore, if each observation in the OC dataset has a group category associated with it, it is possible to check how the patterns are distributed in the latent space with respect to these categories.

In the clinical part of the OC dataset, we focus on a set of relevant features that expert clinicians consider the most relevant. The name and a description of each relevant feature are presented in Table 2. We use a one-hot encoding scheme for categorical variables to construct an autoencoder with input and output layers of size 70. The hidden layer has a size of 3 to allow us to visualize the observations in the latent space, using the hyperbolic tangent function as activation function. The learning process minimizes the mean square error loss function. In Panel (a) of Fig. 1, we show the latent space of the autoencoder with observations of the clinical part of the OC dataset. We split the OC dataset into a training set (points) and a test set (crosses). Observations in blue belong to the platinum-sensitive group, and those in red belong to the platinum-resistant group. A portion of the platinum-sensitive group is separated into a distinct region, but the rest of the observations are mixed between the two groups.

Additionally, we can check the autoencoder latent space representation for classification between platinum-sensitive and platinum-resistant groups if we attach a logistic regression layer (Bishop, 2006) to the encoder substructure of the trained autoencoder. We train this new architecture with the encoder weights and biases frozen and update only those belonging to the logistic regression layer. We can subsequently improve the classification capacity of this new architecture by retraining it with the encoder weights and biases unfrozen, a process called fine tuning (Goodfellow et al., 2016). The latent space of the fine-tuned encoder substructure with the observation of the clinical part of the OC dataset is shown in Panel (b) of Fig. 1. In this case, after the supervised task, the platinum-sensitive and platinum-resistant groups are more defined.

We repeat the same process for the genetic part of the OC dataset using a set of relevant features that expert clinicians considered the most relevant (Table 1). In this part, we have an autoencoder with input and output layers of size 329 after coding categorical variables with a one-hot encoding scheme. Additionally, the hidden layer with hyperbolic tangent activation functions has a size of 3 to enable us to visualize the observations in the latent space, and the learning process minimizes the mean square error loss function. We show the latent space of the autoencoder with the observations of the genetic part of the OC dataset in Panel (c) of Fig. 1. We separate the OC dataset into a training set (points) and a test set (crosses), with observations in blue belonging to the platinum-sensitive group and observations in red belonging to the platinum-resistant group. In this case, observations in both groups are very mixed. As in the clinical part, we can check the autoencoder latent space representation for classification between platinum-sensitive and platinum-resistant groups, namely, we attach a logistic regression layer to the encoder substructure of the trained autoencoder and train it with the encoder weights and biases frozen while updating those belonging to the logistic regression layer. We can increase the classification capacity of this new architecture by fine tuning. The latent space of the fine-tuned encoder substructure with the observations of the genetic part of the OC dataset is shown in Panel (d) of Fig. 1. As in the clinical part, after the supervised task, the platinum-sensitive and platinum-resistant groups are more defined, but they are still very mixed.

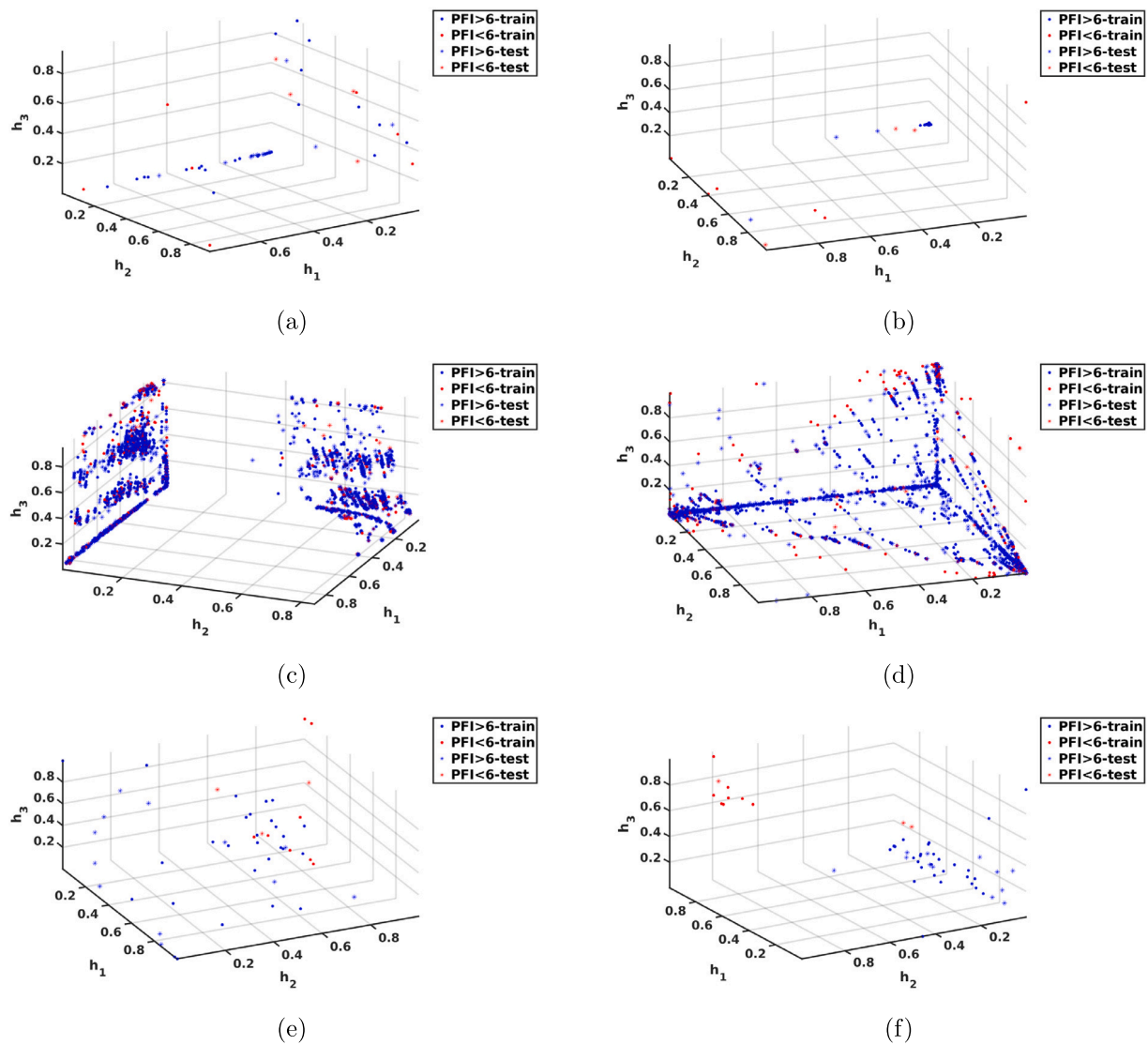

**Fig. 1.** 3-D latent space of an autoencoder for the clinical part of the OC dataset (a) and for the clinical part of the OC dataset after being fine tuned (b). 3-D latent space of an autoencoder for the genetic part of the OC dataset (c) and for the genetic part of the OC dataset after being fine tuned (d). 3-D latent space of an autoencoder for the combination of genes and clinical features of the OC dataset (e) and for the combination of genes and clinical features of the OC dataset after being fine tuned (f).

Finally, we repeat the above process for joining the clinical and genetic part of the OC dataset. In this case, for each patient, we have combined the clinical variables and the genetic profile, indicating genes with mutations for that patient, coding these variables with a one-hot encoding scheme. We show the latent space of the autoencoder with the observations of the combination of genes and clinical features of the OC dataset in Panel (e) of Fig. 1. Also, in Panel (f) of Fig. 1, we can observe the latent space of the fine-tuned encoder substructure with the observations of the combination of genes and clinical features of the OC dataset. In the first one, observations in platinum-sensitive group (blue) and platinum-resistant (red) are mixed. However, in the second one, after the supervised task, the platinum-sensitive and platinum-resistant groups are almost completely separated.

### 4.2. Adaptation of IVI in the OC dataset

In addition to the analyzes of the OC dataset with autoencoders, with the aim of understanding the clinical and genetic dataset in relation to OC progression, we are also interested in studying which features are most important for disease progression classification and the relationships between these features. For this purpose, we used the adaptation of the IVI method presented in Section 3.3. As we did with the autoencoders, we analyzed the set of relevant clinical and genetic features that expert clinicians considered the most relevant (Tables 1 and 2), with categorical features coded with the one-hot encoding scheme. Furthermore, as we have previously seen, each observation of the clinical and genetic parts of the OC dataset is labeled as platinum-resistant or platinum-sensitive, according to the PFI feature.

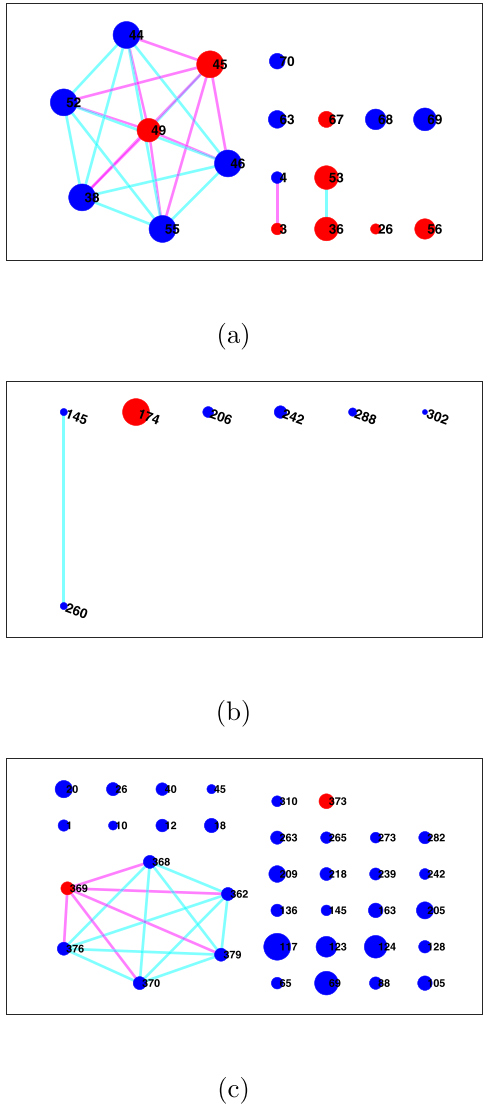

(a)

(b)

(c)

**Fig. 2.** Graph of informative features and their relationships for the clinical part (a), for the genetic part (b), and for combination of genes and clinical features (c) of the OC dataset as a result of using the adaptation of the IVI.

In Panel (a) of Fig. 2 we can see the graph of the informative features and their relationships for the clinical part of OC dataset as result of using the adaptation of the IVI. In this graph, red nodes correspond to negative weights, that is, features that are influential for the platinum-resistant class, and blue nodes correspond to positive weights, indicating features that are influential for the platinum-sensitive class. Similarly, magenta links have negative weights and cyan links have positive weights.

There is a large group of 7 features that are each linked to all the others. These features are the type of surgery (*Surgery*) with *Primary* (38) category; chemotherapy treatment prior to primary surgery (*Neoadjuvance*) with both *No* (44) and *Yes* (45) categories; observed response to neoadjuvant chemotherapy (*Response_of_Neoadjuvance*) with *Not_Applied* (46) and *RP* (49) categories, which indicate that neoadjuvant treatment has not been applied and that after neoadjuvant treatment, there is a presence of detectable cancer, respectively; decision after neoadjuvant treatment (*Attitude_of_Interval_Surgery*) with *Not_Applied* (52) category; and type of interval surgery (*Type_of_Interval_Surgery*), also the *Not_Applied* (55) category. Features *Neoadjuvance: Yes* (45) and *Response_of_Neoadjuvance: RP* (49) are red, so they have negative weights; therefore, they are influential for the platinum-resistant class. All the other features in the group are blue and have positive weights, so they are influential for the platinum-sensitive class.

Furthermore, there are 2 groups of 2 features. The first group consists of information about the presence of gynecological cancer in the family medical history (*Gynecological_Family_History*), with both *No* (3) and *Yes* (4) categories. The first feature is influential for the platinum-resistant class (red), while the second feature is influential for the platinum-sensitive class (blue). The second group is formed by the type of surgery (*Surgery*) with *Interval* (36) category and the decision after neoadjuvant treatment (*Attitude_of_Interval_Surgery*) with *Yes* (53) category, which implies the logical continuation of the disease treatment in a two-step procedure (neoadjuvancy plus interval surgery). Both features are red, indicating that they are influential for the platinum-resistant class.

Finally, we have several independent features. Those influential for the platinum-resistant class (red) are the existence of vascular or perineural invasion (*Perineural_Vascular_Invasion*), *Yes* (26) category; type of interval surgery (*Type_of_Interval_Surgery*), *R0* (56) category, which means a complete resection of the tumor during surgical procedures; and the age at diagnosis (*Age_at_Diagnosis*) (67). The features that are influential for the platinum-sensitive class (blue) are the observed response to adjuvant chemotherapy (*Response_of_Adjuvance*), *RC* category (63), meaning absence of all detectable cancer after treatment administration; progression-free survival (*PFS*) (68), which represents the time from the first date of pharmacological treatment until radiological or biochemical progression; the number of cycles of chemotherapy received in adjuvant therapy (*Cycles_of_Adjuvance*) (69); and the overall survival (*OS*) (70), which estimates the duration of patient survival from the date of diagnosis or treatment initiation.

Panel (b) of Fig. 2 presents the graph for the informative features and their relationships for the genetic part of the OC dataset as a result of using the adaptation of the IVI. As in the previous graph, nodes with negative weights are in red (they are influential for the platinum-resistant class), and nodes with positive weights are in blue (they are influential for the platinum-sensitive class). Moreover, magenta links have negative weights, and cyan links have positive weights.

In this case, there is only one group of 2 features and 5 independent features. The group is composed of the feature that represents the genetic changes from the reference allele to the variant allele (*Genetic_Change*), represented by the *GTGGTGAAGAACATTCAGGCAA>G* (145) category, and the amino acid changes from the reference allele to the variant allele (*Amino_Acids*), represented by the *LPECSSP>-* (260) category. Both features are blue and have positive weights, so they are influential for the platinum-sensitive class. Thus, the variant allele (145) and the subsequent amino acid change (260) are predicted to have a moderate impact on the functionality of BARD1, a nuclear partner of BRCA1. This finding is in accordance with the fact that alterations impairing the activity of the homologous recombination (HR) pathway could behave as low/moderate risk factors in the context of gynecological tumors.

Moreover, the independent features that are influential for the platinum-sensitive class (blue) are amino acid changes from the reference allele to the variant allele (*Amino_Acids*), represented by the *D>G* (206), *I>T* (242), *R>I* (288), and *S>R* (302) categories. All these amino acid changes are associated with genes linked to DNA repair pathways *TP53*, *RAD50*, *ATM*, *FANCM*, *FANCB*, *FANCL* and *SLX4*. Notably, our data science study focusing on text analytics also pinpointed a correlation of *SLX4* alterations with the clinical course of ovarian carcinomas (Bote-Curiel et al., 2021b). On the other hand, the independent feature that is influential for the platinum-resistant class (red) is the genotype *Genotype, UNC_Homo* category. The association of such genetic features with chemotherapy resistance could potentially be explained by the loss of wild-type alleles (loss of heterozygosity) or

**Table 3**

Encoders and classifier with the number of features used in the clinical part of the OC dataset and the accuracy and AUC in the test set.

| Classifier | Number of variables | Testing accuracy | Testing AUC |
|---|---|---|---|
| 3-D latent space of autoencoder + logistic regression | 70 (all the variables) | 85.19% | 0.74 |
| 3-D latent space of fine-tuned autoencoder + logistic regression | 70 (all the variables) | 72.22% | 0.66 |
| 3-D latent space of autoencoder + logistic regression | 18 (a set of variables chosen by adapted IVI) | 87.04% | 0.75 |
| 3-D latent space of fine-tuned autoencoder + logistic regression | 18 (a set of variables chosen by adapted IVI) | 88.89% | 0.76 |
| 10-D latent space of autoencoder + logistic regression | 70 (all the variables) | 79.63% | 0.64 |
| 10-D latent space of fine-tuned autoencoder + logistic regression | 70 (all the variables) | 74.07% | 0.60 |
| 10-D latent space of autoencoder + logistic regression | 18 (a set of variables chosen by adapted IVI) | 83.33% | 0.72 |
| 10-D latent space of fine-tuned autoencoder + logistic regression | 18 (a set of variables chosen by adapted IVI) | 87.03% | 0.78 |
| 20-D latent space of autoencoder + logistic regression | 70 (all the variables) | 75.93% | 0.61 |
| 20-D latent space of fine-tuned autoencoder + logistic regression | 70 (all the variables) | 75.93% | 0.61 |
| 20-D latent space of autoencoder + logistic regression | 18 (a set of variables chosen by adapted IVI) | 88.89% | 0.79 |
| 20-D latent space of fine-tuned autoencoder + logistic regression | 18 (a set of variables chosen by adapted IVI) | 87.04% | 0.75 |

**Table 4**

Encoders and classifier with the number of features used in the combination of genes and clinical features of the OC dataset and the accuracy and AUC in the test set.

| Classifier | Number of variables | Testing accuracy | Testing AUC |
|---|---|---|---|
| 3-D latent space of autoencoder + logistic regression | 395 (all the variables) | 76.92% | 0.62 |
| 3-D latent space of fine-tuned autoencoder + logistic regression | 395 (all the variables) | 84.62% | 0.67 |
| 3-D latent space of autoencoder + logistic regression | 36 (a set of variables chosen by adapted IVI) | 88.46% | 0.83 |
| 3-D latent space of fine-tuned autoencoder + logistic regression | 36 (a set of variables chosen by adapted IVI) | 86.54% | 0.81 |
| 10-D latent space of autoencoder + logistic regression | 395 (all the variables) | 80.77% | 0.65 |
| 10-D latent space of fine-tuned autoencoder + logistic regression | 395 (all the variables) | 80.77% | 0.65 |
| 10-D latent space of autoencoder + logistic regression | 36 (a set of variables chosen by adapted IVI) | 84.62% | 0.84 |
| 10-D latent space of fine-tuned autoencoder + logistic regression | 36 (a set of variables chosen by adapted IVI) | 82.69% | 0.79 |
| 20-D latent space of autoencoder + logistic regression | 395 (all the variables) | 78.85% | 0.56 |
| 20-D latent space of fine-tuned autoencoder + logistic regression | 395 (all the variables) | 82.69% | 0.66 |
| 20-D latent space of autoencoder + logistic regression | 36 (a set of variables chosen by adapted IVI) | 76.92% | 0.79 |
| 20-D latent space of fine-tuned autoencoder + logistic regression | 36 (a set of variables chosen by adapted IVI) | 76.92% | 0.75 |

genetic variant gains/amplifications in the more advanced and polytreated tumors and, therefore, the classification of such variants as homozygous in the sequencing studies we performed.

Finally, Panel (c) of Fig. 2 presents the graph for the informative features and their relationships for the combination of genes and clinical variables belonging to the OC dataset as a result of using the adaptation of the IVI. In this case, there is a pattern similar to the clinical part where some clinical variables are related to each other. For example, we have relations among chemotherapy treatment prior to primary surgery (*Neoadjuvance*) with both *No* (368) and *Yes* (369) categories, the type of surgery (*Surgery*) with *Primary* (362) category, the decision after neoadjuvant treatment (*Attitude_of_Interval_Surgery*) (376) category, the observed response to neoadjuvant chemotherapy (*Response_of_Adjuvance*) (370), and the type of interval surgery (*Type_of_Interval_Surgery*) (379). All of them except one are in blue, indicating that they are influential for the platinum-sensitive class. As an independent variable, the variable *Response_of_Neoadjuvance: RP* (373) is in red, so it has negative weight and it is influential for the platinum-resistant class. The rest of independent variables are in blue, so they are influential for the platinum-sensitive class. For example, we have the gen *APC* with mutation *A>G* (1), the gen *ATM* with mutation *C>G* (10), or the *SLX* with mutation *A>C* (263), among others.

In addition to these results, we have used the subset of features selected by the adaptation of the IVI algorithm with several classifiers and architectures in the task of classifying the observations into the platinum-sensitive and platinum-resistant groups for the clinical part of the OC dataset and the combination of genes and clinical features. In particular, in the clinical part of the OC dataset, we have used the subset of features selected (18 features) with logistic regression, with support-vector machine (SVM), with decision tree, and with k-nearest neighbors (KNN) classifiers, and also with several encoder structures, using said trained autoencoders with different bottleneck sizes (3, 10, and 20) subsequently followed by a logistic regression layer. These encoder structures were first trained by updating only the logistic regression layer and later fine tuning the whole structure

**Table 5**

Results of AUC for a test set in the clinical part of the OC dataset for all the features and some sets of feature selected for a several selection methods and classifiers, where LR is for logistic regression, DT for decision tree, SVM for support-vector machine, and KNN for k-nearest neighbors. Also, we show the number of total variables used for each feature extraction method.

| | Num var. | AUC LR | AUC DT | AUC SVM | AUC KNN |
|---|---|---|---|---|---|
| All features | 70 | 0.66 | 0.89 | 0.77 | 0.89 |
| IVI | 18 | 0.78 | 0.89 | 0.94 | 0.90 |
| JMI | 6 | 0.87 | 0.9 | 0.92 | 0.79 |
| CMIM | 7 | 0.87 | 0.9 | 0.9 | 0.93 |
| MIM | 3 | 0.84 | 0.87 | 0.82 | 0.86 |
| MRMR | 6 | 0.83 | 0.89 | 0.83 | 0.85 |
| MIFS | 6 | 0.9 | 0.86 | 0.9 | 0.91 |
| DISR | 4 | 0.79 | 0.87 | 0.84 | 0.84 |
| CIFE | 7 | 0.87 | 0.89 | 0.94 | 0.85 |
| ICAP | 6 | 0.77 | 0.9 | 0.89 | 0.81 |
| CondRed | 4 | 0.92 | 0.87 | 0.95 | 0.76 |
| CMI | 6 | 0.74 | 0.9 | 0.78 | 0.81 |
| Relief | 5 | 0.62 | 0.47 | 0.56 | 0.66 |

was performed. This process was also made using the complete set of clinical features (70 features) as a reference. Table 3 presents some encoders with the logistic regression classifier, the number of features used (complete set or the adaptation of the IVI subset), and their accuracy and the area under the curve (AUC) values in the test set. Also, in Table 5, we have the AUC results for the logistic regression, SVM, decision tree, and KNN classifiers. The free parameters of the SVM and KNN classifiers, that is, the empirical cost parameter $C$ of the SVM and the number of nearest neighbors ($K$) of KNN, were both adjusted by a cross validation (CV) process over a training dataset. Parameter $C$ was selected from the set of values $\{1, 10, 100, 1000\}$ and $K$ from the set of values $\{2, 3, 4, \ldots, 18, 19, 20\}$. These results show that using the adaptation of the IVI subset of features yields better performance (test accuracy and AUC) than using the complete set of features in the classifiers and in the encoders.

We repeated the above process for the genetic part of the OC dataset and for the combination of genes and clinical features. In the first case,

**Table 6**

Results of AUC for a test set for the combination of genes and clinical features of the OC dataset for all the features and some sets of feature selected for a several selection methods and classifiers, where LR is for logistic regression, DT for decision tree, SVM for support-vector machine, and KNN for k-nearest neighbors. Also, we show the number of total variables and the number of genetic variables used for each feature extraction method.

|  | Num var. | Num. genetic var. | AUC LR | AUC DT | AUC SVM | AUC KNN |
|---|---|---|---|---|---|---|
| All features | 395 | 326 | 0.78 | 0.67 | 0.81 | 0.90 |
| IVI | 36 | 29 | 0.80 | 0.70 | 0.85 | 0.80 |
| JMI | 2 | 0 | 0.65 | 0.61 | 0.76 | 0.78 |
| CMIM | 2 | 0 | 0.61 | 0.61 | 0.76 | 0.78 |
| MIM | 2 | 0 | 0.61 | 0.61 | 0.76 | 0.78 |
| MRMR | 3 | 1 | 0.65 | 0.61 | 0.66 | 0.71 |
| MIFS | 3 | 1 | 0.65 | 0.61 | 0.66 | 0.71 |
| DISR | 2 | 1 | 0.61 | 0.63 | 0.76 | 0.71 |
| CIFE | 2 | 1 | 0.72 | 0.71 | 0.74 | 0.71 |
| ICAP | 2 | 1 | 0.70 | 0.67 | 0.79 | 0.71 |
| CondRed | 2 | 0 | 0.72 | 0.61 | 0.76 | 0.78 |
| CMI | 2 | 0 | 0.76 | 0.61 | 0.76 | 0.78 |
| Relief | 16 | 11 | 0.48 | 0.46 | 0.47 | 0.53 |

the complete set of features included 329 features, and the subset of features selected by the adaptation of the IVI algorithm included 7 features. However, the AUC is around 0.5 in all the cases. This is due to fact that the genetic part of the OC dataset is very unbalanced, and the classifiers predict just the majority class for all the data points. In the second case (Tables 4 and 6), we have that the complete set of features are 395 features and the subset of features selected by the adaptation of the IVI algorithm are 36 features. In this case, results reveal that using the adaptation of the IVI subset of features results in the most cases with better performance than using the complete set of features in the classifiers and in the encoders. Moreover, the number of features used is reduced from 395 to 36. Therefore, we can conclude that the features selected in both the clinical and the combination of genes and clinical features are informative enough for the classification of platinum-sensitive and platinum-resistant groups.

As a final experiment, we compared the IVI algorithm with other feature selection methods. To conduct this comparison, we used a series of classifiers to evaluate the performance of the features selected by each method. Specifically, we have *Relief* algorithm (Kononenko, 1994) and a set of methods based on mutual information considerations, namely, MRMR, MIFS, CMIM, ICAP, MIM, DISR, JMI, CIFE, CondRed, and CMI (Brown et al., 2012). However, these methods only returns a ranking of features by importance. To select the most important variables for each ranking, we used the *Forward Selection Step with Mutual Information* technique (Brown et al., 2012). This technique provided us with the number of variables that are most important in each ranking, starting from the first one. We then evaluated the performance of the features selected by each method with logistic regression, SVM, KNN, and decision tree classifiers. In Table 5, we can examine the results of the AUC for a test set for the clinical part of the OC dataset. We can highlight that the performance of IVI is among the best in most classifiers. In the same way, for the combination of genes and clinical features of the OC dataset, we present in Table 6 the performance with IVI and the rest of classifiers. In this case, the performance of IVI is the best in all classifiers by a significant margin. This may be due to the fact that the set of methods based on mutual information does not work well with this type of complex dataset, where all the genes are categorical variables and many of the clinical variables are metric. This can be confirmed by observing in the Table 6 that IVI selects 36 important variables, 29 of which are genetic variables, while most of the other methods select far fewer variables, with almost no consideration of genetic variables. In addition to these results, it should be noted that IVI also provides us with interpretability because it gives the relationships among features. In some application domains, and specially in OC disease management, this can be even more important than performance itself.

## 5. Discussion and conclusion

OC is the second most common gynecological neoplasm, the gynecological tumor with worst prognosis, and the fifth leading cause of cancer-associated mortality. This makes OC one of the most serious health challenges worldwide, with one major reason being the lack of robust predictive and prognostic molecular biomarkers underpinning *a priori* knowledge of the evolution of the disease. Thus, in this work, we have focused on both extensive clinical information collected from patients diagnosed with OC and genomic data obtained through whole-exome or targeted sequencing, with the aim of identifying relationships between clinical and genetic factors and OC disease progression-related variables. We conducted two analyses. First, we used autoencoders, a nonlinear feature extraction method, to assess whether the OC dataset had intrinsic or natural separability patterns between disease progression groups (platinum-resistant and platinum-sensitive groups). The results for the clinical part of the OC dataset show that part of the platinum-sensitive group is separated, but the rest of the observations are mixed between the two groups. However, after fine tuning the autoencoder in a supervised manner, the new latent space of the autoencoder shows a more defined separation of the platinum-sensitive and platinum-resistant groups. Regarding the genetic part of the OC dataset, the observations in both groups are very mixed, exhibiting no separability. Moreover, after supervised training, the new latent space shows that the platinum-sensitive and platinum-resistant groups are more defined, but they are still quite mixed. In the combination of genes and clinical features of the OC dataset, observations are mixed between platinum-sensitive and platinum-resistant groups, but after fine tuning, the new latent space shows separation for both groups.

Then, we employed an adaptation of the IVI method to study which features of the OC dataset are most important for the disease progression classification and to assess the relationships among these features. The results confirm the predictive and prognostic role of certain variables related to the initial treatment of OC. In this setting, interval surgery and the corresponding neoadjuvant treatment are frequent in cases with a higher tumor burden at the time of diagnosis and which consequently require initial chemotherapeutic treatment to reduce this burden. The association that our study shows between variables related to interval surgery and the appearance of therapeutic resistance is consistent with other studies that confirm a longer overall survival in these patients, possibly due to a limited response of certain patients to neoadjuvant treatment.

In terms of genetic factors, many previous studies, both genomic and clinical, have demonstrated a clear association between the presence of alterations in genes related to DNA repair pathways (BRCA1/2, RAD51C and other loci that act as low/moderate risk factors) and the clinical course of the disease. Despite the limited series of patients that have been sequenced in this study, the selection of extreme cases in terms of response to chemotherapy (high responders vs. cases with intrinsic resistance) and the application of useful data science strategies have made it possible to define certain correlations with genes involved in these molecular pathways. It is conceivable that by expanding the series of OC cases, restricting the histopathologies included in the sequencing, developing databases that interrelate multiple -omics data from each patient and applying more refined analysis tools we will be able to define, in the medium-term future, combinations of biomarkers that more efficiently predict the response to therapies and the clinical course of this disease.

On the one hand, the analysis of the available OC dataset with autoencoders enables us to evaluate the implicit information contained in the clinical and genetic data in such a way that we can scrutinize how these data are distributed with respect to the disease progression groups. On the other hand, the analysis of the dataset with the adaptation of the IVI has provided explicit information and knowledge on which clinical and genetic features are most informative regarding OC disease progression classification. These analyses complement the two previous works that paid specific attention to the analysis of the OC dataset with univariate and linear multivariate methods, respectively (Bote-Curiel et al., 2021a, 2021b). All these analyses together have enabled us to obtain very detailed information from the OC dataset and together represent an approach to obtain as much information as possible from clinical data from a principled and detailed statistical and machine learning analysis.

As a limitation, we scrutinized the changes in performance between the complete set of features and the subset of features selected by IVI for the classifiers analyzed, and not statistically significant differences could be consistently determined even when using several different non-parametric tests. Analysis of surrogate data showed that this can be due to the standard error being large due to the small size of the clinical patient sample of the OC dataset. Nevertheless, the trends were clear to IVI providing in general a reduction in the number of features while maintaining the accuracy. In the future, more patients and some other types of genetic or multiomic data may be included in the studies. However, these data are often expensive to obtain and it is necessary to have techniques capable of assembling and analyzing all the heterogeneous data types together. Therefore, it is highly relevant to have statistics-based machine learning tools that can provide as much information as possible while being as interpretable as possible in future multiomic scenarios. Also, in future works, we plan to extend the functioning of IVI algorithm for the detection of nonlinear relationships between features. To do this, we can use new correlation coefficients that capture nonlinear dependency between features or use nonlinear machine learning weights. However, to find out which linear and non-linear methods work best in IVI, we should do an in-depth algorithmic study.

**CRediT authorship contribution statement**

**Luis Bote-Curiel:** Designed and organized the paper, Performed the data analysis, Wrote the methods and part of the introduction, Results, Conclusion. **Sergio Ruiz-Llorente:** Wrote the dataset description and part of the introduction, Results, Conclusion. **Sergio Muñoz-Romero:** Designed and organized the paper, Writing some parts and reviewing the manuscript. **Mónica Yagüe-Fernández:** Provided the data processing. **Arantzazu Barquín:** Provided the data processing. **Jesús García-Donas:** Writing some parts and reviewing the manuscript. **José Luis Rojo-Álvarez:** Designed and organized the paper, Writing some parts and reviewing the manuscript.

**Declaration of competing interest**

The authors declare that they have no known competing financial interests or personal relationships that could have appeared to influence the work reported in this paper.

**Acknowledgments**

This work has been supported by the meHeart-RisBi and Beyond grants (PID2019-104356RB-C42 and PID2019-106623RB-C41) from the Spanish Ministry of Science and Innovation and by the SC-LEARNING-CM grant supported by the REACT-EU programme (Next Generation EU funds) from the Community of Madrid and Rey Juan Carlos University, spain. All authors read and approved the final manuscript.

**Appendix. Synthetic examples**

*A.1. Autoencoder synthetic example*

To illustrate the functioning of autoencoders, we present a simple example using a synthetic dataset. We first generate the synthetic dataset composed of 1000 individuals and 6 mixed variables (categorical and metric). There is one group of 2 dependent metric variables, another group of 2 dependent categorical variables (each with 2 categories), and 2 completely independent variables, one metric and one categorical (with 2 categories). One-hot encoding is used for the categorical variables; thus, the architecture of the autoencoder has an input and output layer size of 9. The hidden layer has a size of 3 to enable us to visualize the observations in the latent space. In this case, the learning process minimizes the mean square error loss function. In Panel (a) of Fig. A.3, we show the observations in the latent space, where points represent the training set and crosses represent the test set, with category 1 in red and category 0 in blue.

*A.2. Adaptation of IVI synthetic example*

We propose a simple example to illustrate how the adaptation of the IVI method works. We generated a dataset similar to the one used in the autoencoder examples, that is, a dataset composed of 1000 observations and 6 mixed variables. There is one group of 2 categorical variables (each with 2 categories), another group of 2 dependent metric variables, and 2 completely independent variables, one categorical (with 2 categories) and one metric. After applying the one-hot encoding scheme for the categorical variables, the dataset has dimensions of 1000 by 9. Additionally, the variables in each group are highly correlated but are weakly correlated with out-group variables. In addition, variables in first group, variables in second group, the first independent variable, and the second independent variable have a correlation with the response variable (binary variable) of 0.7, 0.1, 0.5, and 0.1, respectively.

In Panel (b) of Fig. A.3, we show the graph resulting from the application of the adaptation of the IVI to the synthetic dataset. In the graph, red nodes have negative weights (influential for class 0) and blue nodes have positive weights (influential for class 1). Similarly, magenta links have negative weights and cyan links have positive weights. In the figure, there is a group of four variables that are all connected to each other. This group corresponds to the first group of two categorical variables with two categories each. Variable 1 and variable 2 are complementary (two categories), and variable 3 and variable 4 are also complementary (two categories). This group of variables is highly correlated with each other and also with the response variable; therefore, they are identified as relevant. The second group that we observe is the first independent categorical variable, which has 2 categories that are linked and complementary.

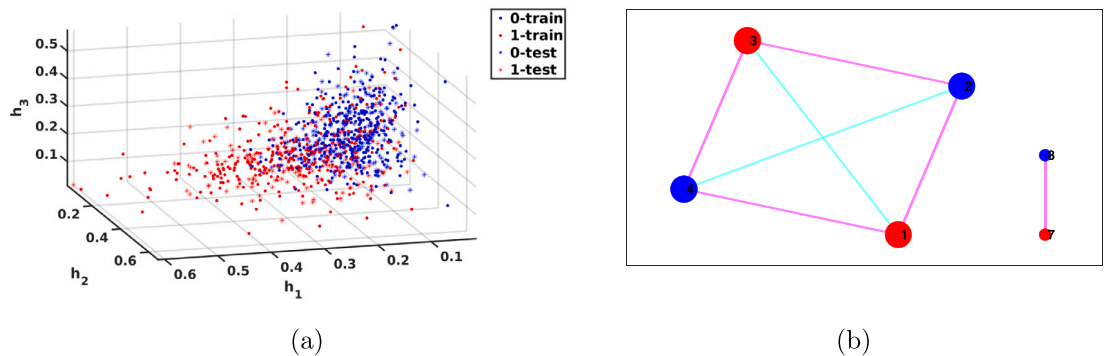

**Fig. A.3.** 3-D latent space of an autoencoder for an example using synthetic data (a). Graph of the informative features and their relationships for a simple example with synthetic data as a result of using the adaptation of the IVI method (b).

## References

Alipoor, M., Khani Parashkoh, M., & Haddadnia, J. (2010). A novel biomarker discovery method on protemic data for ovarian cancer classification. In *2010 18th iranian conference on electrical engineering* (pp. 1–6).

Battiti, R. (1994). Using mutual information for selecting features in supervised neural net learning. *IEEE Transactions on Neural Networks*, *5*(4), 537–550.

Bennasar, M., Hicks, Y., & Setchi, R. (2015). Feature selection using joint mutual information maximisation. *Expert Systems with Applications*, *42*(22), 8520–8532.

Bishop, C. (2006). *Pattern recognition and machine learning* (1st ed.). New York, NY, US: Springer-Verlag.

Bote-Curiel, L., Ruiz, S., Muñoz-Romero, S., Yagüe-Fernández, M., Barquín, A., García-Donás, J., et al. (2021a). A resampling univariate analysis approach to ovarian cancer from clinical and genetic data. *IEEE Access*, *9*, 25959–25972.

Bote-Curiel, L., Ruiz, S., Muñoz-Romero, S., Yagüe-Fernández, M., Barquín, A., García-Donás, J., et al. (2021b). Text analytics and mixed feature extraction in Ovarian cancer clinical and genetic data. *IEEE Access*, *9*, 58034–58051.

Brown, G., Pocock, A., Zhao, M.-J., & Luján, M. (2012). Conditional likelihood maximisation: A unifying framework for information theoretic feature selection. *Journal of Machine Learning Research*, *13*(2), 27–66.

Ding, C., & Peng, H. (2005). Minimum redundancy feature selection from microarray gene expression data. *Journal of Bioinformatics and Computational Biology*, *03*(02), 185–205.

Efron, B., & Tibshirani, R. (1986). Bootstrap methods for standard errors, confidence intervals, and other measures of statistical accuracy. *Statistical Science*, *1*(1), 54–75.

Elias, K. M., Fendler, W., Stawiski, K., Fiascone, S. J., Vitonis, A. F., Berkowitz, R. S., et al. (2017). Diagnostic potential for a serum miRNA neural network for detection of ovarian cancer. *ELife*, *6*, Article e28932.

Fleuret, F. (2004). Fast binary feature selection with conditional mutual information. *Journal of Machine Learning Research*, *5*, 1531–1555.

Goodfellow, I., Bengio, Y., & Courville, A. (2016). *Deep learning*. Cambridge, MA, US: The MIT Press.

Hinton, G. E., & Salakhutdinov, R. R. (2006). Reducing the dimensionality of data with neural networks. *Science*, *313*(5786), 504–507.

Jakulin, A. (2005). *Machine learning based on attribute interactions*. (Ph.D. thesis), Faculty of Computer and Information Science, University of Ljubljana.

Kira, K., & Rendell, L. A. (1992). A practical approach to feature selection. In D. Sleeman, & P. Edwards (Eds.), *Machine learning proceedings 1992* (pp. 249–256). San Francisco (CA): Morgan Kaufmann.

Kononenko, I. (1994). Estimating attributes: Analysis and extensions of RELIEF. In F. Bergadano, & L. De Raedt (Eds.), *Machine learning: ECML-94* (pp. 171–182). Berlin, Heidelberg: Springer Berlin Heidelberg.

Lewis, D. D. (1992). Feature selection and feature extraction for text categorization. In *Speech and natural language: proceedings of a workshop held at harriman*.

Liang, J., Hou, L., Luan, Z., & Huang, W. (2019). Feature selection based on conditional mutual information considering feature interaction. *Symmetry*, *11*(7).

Lin, D., & Tang, X. (2006). Conditional infomax learning: An integrated framework for feature extraction and fusion. In A. Leonardis, H. Bischof, & A. Pinz (Eds.), *Computer vision – ECCV 2006* (pp. 68–82). Berlin, Heidelberg: Springer Berlin Heidelberg.

Macintyre, G., Goranova, T. E., De Silva, D., Ennis, D., Piskorz, A. M., Eldridge, M., et al. (2018). Copy number signatures and mutational processes in Ovarian Carcinoma. *Nature Genetics*, *50*(9), 1262–1270.

Martínez-Ramón, M., Gupta, A., Rojo-Álvarez, J. L., & Christodoulou, C. (2020). *Machine learning applications in electromagnetics and antenna array processing* (1st ed.). London, UK: Artech House UK.

Meyer, P. E., & Bontempi, G. (2006). On the use of variable complementary for feature selection in cancer classification. In *EuroGP'06, Proceedings of the 2006 international conference on applications of evolutionary computing* (pp. 91—102). Berlin, Heidelberg: Springer-Verlag.

Moschetta, M., George, A., Kaye, S. B., & Banerjee, S. (2016). BRCA somatic mutations and epigenetic BRCA modifications in serous Ovarian cancer. *Annals of Oncology*, *27*(8), 1449–1455.

Muñoz-Romero, S., Gorostiaga, A., Soguero-Ruiz, C., Mora-Jiménez, I., & Rojo-Álvarez, J. L. (2020). Informative variable identifier: Expanding interpretability in feature selection. *Pattern Recognition*, *98*, Article 107077.

Peng, H., Long, F., & Ding, C. (2005). Feature selection based on mutual information criteria of max-dependency, max-relevance, and min-redundancy. *IEEE Transactions on Pattern Analysis and Machine Intelligence*, *27*(8), 1226–1238.

Pujade-Lauraine, E., & Combe, P. (2016). Recurrent ovarian cancer. *Annals of Oncology*, *27*, i63–i65.

Radovic, M., Ghalwash, M., Filipovic, N., & Obradovic, Z. (2017). Minimum redundancy maximum relevance feature selection approach for temporal gene expression data. *BMC Bioinformatics*, *18*(1), 9.

Ramírez-Gallego, S., Lastra, I., Martínez-Rego, D., Bolón-Canedo, V., Benítez, J. M., Herrera, F., et al. (2017). Fast-mRMR: Fast minimum redundancy maximum relevance algorithm for high-dimensional big data. *International Journal of Intelligent Systems*, *32*(2), 134–152.

Remeseiro, B., & Bolon-Canedo, V. (2019). A review of feature selection methods in medical applications. *Computers in Biology and Medicine*, *112*, Article 103375.

Saeys, Y., Inza, I. n., & Larrañaga, P. (2007). A review of feature selection techniques in bioinformatics. *Bioinformatics*, *23*(19), 2507–2517.

Song, H.-J., Yang, E.-S., Kim, J.-D., Park, C.-Y., Kyung, M.-S., & Kim, Y.-S. (2018). Best serum biomarker combination for ovarian cancer classification. *BioMedical Engineering OnLine*, *17*(152).

Stewart, C., Ralyea, C., & Lockwood, S. (2019). Ovarian cancer: An integrated review. *Seminars in Oncology Nursing*, *35*(2), 151–156.

Tadist, K., Najah, S., Nikolov, N. S., Mrabti, F., & Zahi, A. (2019). Feature selection methods and genomic big data: a systematic review. *Journal of Big Data*, *6*(1), 79.

Urbanowicz, R. J., Meeker, M., La Cava, W., Olson, R. S., & Moore, J. H. (2018). Relief-based feature selection: Introduction and review. *Journal of Biomedical Informatics*, *85*, 189–203.

Vergara, J. R., & Estévez, P. A. (2014). A review of feature selection methods based on mutual information. *Neural Computing and Applications*, *24*(1), 175–186.

Yang, H. H., & Moody, J. (1999). Data visualization and feature selection: New algorithms for nongaussian data. In *NIPS'99, Proceedings of the 12th international conference on neural information processing systems* (pp. 687—693). Cambridge, MA, USA: MIT Press.

Yu, K., Hu, V., Wang, F., Matulonis, U. A., Mutter, G. L., Golden, J. A., et al. (2020). Deciphering serous ovarian carcinoma histopathology and platinum response by convolutional neural networks. *BMC Medicine*, *18*(1), 236.

Zhu, W., Xie, L., Han, J., & Guo, X. (2020). The application of deep learning in cancer prognosis prediction. *Cancers*, *12*(3), 603.